\documentclass[10pt,twocolumn,letterpaper]{article}

\usepackage{cvpr}
\usepackage{times}
\usepackage{epsfig}
\usepackage{graphicx}
\usepackage{amsmath}
\usepackage{amssymb}

\usepackage{subcaption}
\usepackage{multicol}
\usepackage{algorithm}
\usepackage[noend]{algpseudocode}
\makeatletter
\def\BState{\State\hskip-\ALG@thistlm}
\makeatother

\usepackage[breaklinks=true,bookmarks=false]{hyperref}

\cvprfinalcopy 

\def\cvprPaperID{****} 

\begin{document}

\title{DSSLIC: Deep Semantic Segmentation-based Layered Image Compression}

\author{Mohammad Akbari, Jie Liang\\
School of Engineering Science, Simon Fraser University, Canada\\
{\tt\small akbari@sfu.ca, jiel@sfu.ca}
\thanks{This work is supported by Google Chrome University Research program and the Natural Sciences and Engineering Research Council (NSERC) of Canada under grant RGPIN312262 and RGPAS478109.}
\and
Jingning Han\\
Google Inc.\\
{\tt\small jingning@google.com}
}

\maketitle

\begin{figure*}
\begin{subfigure}[b]{1\linewidth}
 \centering
  \centerline{\includegraphics[width=\textwidth]{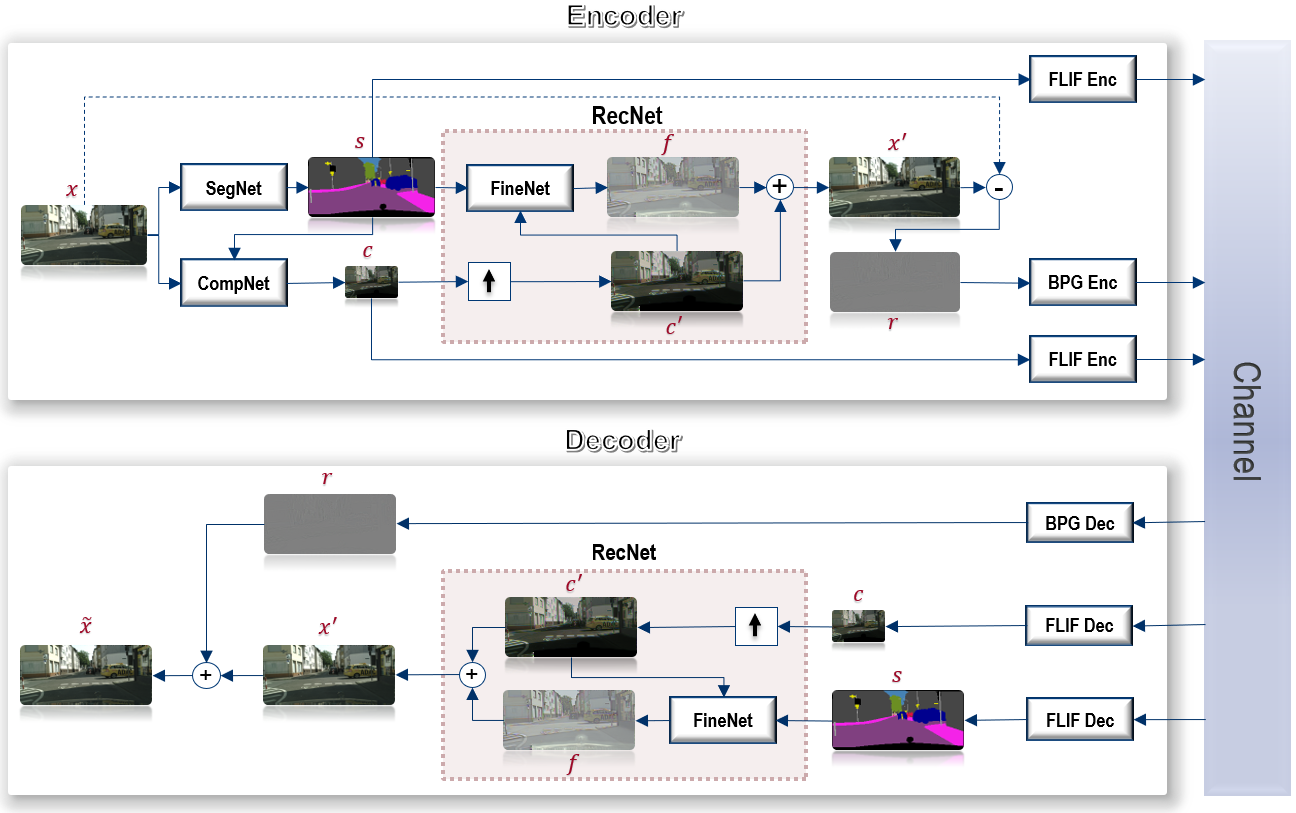}}
\end{subfigure}
\caption{The overall framework of the proposed deep semantic segmentation-based layered image compression (DSSLIC) codec.}
\label{fig:framework}
\end{figure*}

\begin{abstract}
Deep learning has revolutionized many computer vision fields in the last few years, including learning-based image compression. In this paper, we propose a deep semantic segmentation-based layered image compression (DSSLIC) framework in which the semantic segmentation map of the input image is obtained and encoded as the base layer of the bit-stream. A compact representation of the input image is also generated and encoded as the first enhancement layer. The segmentation map and the compact version of the image are then employed to obtain a coarse reconstruction of the image. The residual between the input and the coarse reconstruction is additionally encoded as another enhancement layer. Experimental results show that the proposed framework outperforms the H.265/HEVC-based BPG and other codecs in both PSNR and MS-SSIM metrics across a wide range of bit rates in RGB domain. Besides, since semantic segmentation map is included in the bit-stream, the proposed scheme can facilitate many other tasks such as image search and object-based adaptive image compression\footnote{The source code of the paper: \url{https://github.com/makbari7/DSSLIC}}.
\end{abstract}

\section{Introduction}
\label{Introduction}

Since 2012, deep learning has revolutionized many computer vision fields such as image classification, object detection, and face recognition. In the last couple of years, it has also made some impacts to the well-studied topic of image compression, and in some cases has achieved better performance than JPEG2000 and the H.265/HEVC-based BPG image codec \cite{agustsson2017soft,Agustsson18-2,balle2016end,johnston2017improved,rippel2017real,Santurkar17,theis2017lossy,toderici2015variable,toderici2017full}, making it a very promising tool for the next-generation image compression.

One advantage of deep learning is that it can extract much more accurate semantic segmentation map from a given image than traditional methods \cite{zhao2017pyramid}. Recently, it was further shown that deep learning can even synthesize a high-quality and high-resolution image using only a semantic segmentation map as input \cite{wang2017high}, thanks to the generative adversarial networks (GAN) \cite{goodfellow2014generative}. This suggests the possibility of developing efficient image compression using deep learning-based semantic segmentation and the associated image synthesis.

GAN architecture is composed of two networks named discriminator and generator, which are trained at the same time \cite{goodfellow2014generative}. The generator model $G(z)$ captures the data distribution by mapping the latent $z$ to data space, while the discriminator model $D(x) \in [0, 1]$ estimates the probability that $x$ is a real training sample or a fake sample synthesized by $G$. These two models compete in a two-player minimax game in which the objective function is to find a binary classifier $D$ that discriminates the real data from the fake (generated) ones and simultaneously encourages $G$ to fit the true data distribution. This goal is achieved by minimizing/maximizing the binary cross entropy: 
\begin{equation}
\mathcal{L}_{GAN} =\mathbb{E}_{x\sim p_{data}(x)}[\log D(x)] + \mathbb{E}_{z\sim p_{z}(z)}[\log(1 - D(G(z)))]
\end{equation}
where $G$ tries to minimize this objective against $D$ that tries to maximize it. 

In this paper, we employ GAN to propose a deep semantic segmentation-based layered image compression (DSSLIC) framework as shown in Figure \ref{fig:framework}. In our approach, the semantic segmentation map of the input image is extracted by a deep learning network and losslessly encoded as the base layer of the bit-stream. Next, the input image and the segmentation map are used by another deep network to obtain a low-dimensional compact representation of the input, which is encoded into the bit-stream as the first enhancement layer. After that, the compact image and the segmentation map are used to obtain a coarse reconstruction of the image. The residual between the input and the coarse reconstruction is encoded as the second enhancement layer in the bit-stream. To improve the quality, the synthesized image from the segmentation map is designed to be a residual itself, which aims to compensate the difference between the upsampled version of the compact image and the input image. Therefore the proposed scheme includes three layers of information.

Experimental results in the RGB (4:4:4) domain show that the proposed framework outperforms the H.265/HEVC-based BPG codec \cite{bellard2017bpg} in both PSNR and multi-scale structural similarity index (MS-SSIM) \cite{wang2003multiscale} metrics across a large range of bit rates, and is much better than JPEG, JPEG2000, and WebP \cite{webp2018}. For example, our method can be 4.7 dB better in PSNR than BPG for some Kodak testing images. Moreover, since semantic segmentation map is included in the bit-stream, the proposed scheme can facilitate many other tasks such as image search and object-based adaptive image compression.

The idea of semantic segmentation-based compression was already studied in MPEG-4 object-based video coding in the 1990's \cite{Talluri97}. However, due to the lack of high-quality and fast segmentation methods, object-based image/video coding has not been widely adopted. Thanks to the rapid development of deep learning algorithms and hardware, it is now the time to revisit this approach.

This paper is organized as follows. In Section \ref{Related Works}, the works related to learning-based image compression are briefly reviewed. The architecture of the proposed framework and the corresponding formulation and objective functions are described in Section \ref{Proposed Model}. In Section \ref{Experiments}, the performance of the proposed method is evaluated and compared with the JPEG, JPEG2000, WebP, and BPG codecs.


\section{Related Works}
\label{Related Works}

In traditional compression methods, many components such as entropy coding are hand-crafted. Deep learning-based approaches have the potential of automatically discovering and exploiting the features of the data; thereby achieving better compression performance.

In the last few years, various learning-based image compression frameworks have been proposed. In \cite{toderici2015variable,toderici2017full}, long short-term memory (LSTM)-based recurrent neural networks (RNNs) were used to extract binary representations, which were then compressed with entropy coding. Probability estimation in the entropy coding was also handled by LSTM convolution. Johnston et al. \cite{johnston2017improved} utilized structural similarity (SSIM) loss \cite{wang2004image} and spatially adaptive bit allocation to further improve the performance. 

In \cite{balle2016end}, a scheme that involved a generalized divisive normalization (GDN)-based nonlinear analysis transform, a uniform quantizer, and a nonlinear synthesis transform were developed. Theis et al. \cite{theis2017lossy} proposed a compressive autoencoder (AE) where the quantization was replaced by a smooth approximation, and a scaling approach was used to get different rates. In \cite{agustsson2017soft}, a soft-to-hard vector quantization approach was introduced, and a unified formulation was developed for both the compression of deep learning models and image compression.
 
GAN has been exploited in a number of learning-based image compression schemes. In \cite{Santurkar17}, a discriminator was used to help training the decoder. A perceptual loss based on the feature map of an ImageNet-pretrained AlexNet was introduced although only low-resolution image coding results were reported in \cite{Santurkar17}. In \cite{rippel2017real}, AE was embedded in the GAN framework in which the feature extraction adopted pyramid and interscale alignment. The discriminator also extracted outputs from different layers, similar to the pyramid feature generation. An adaptive training was used where the discriminator was trained and a confusion signal was propagated through the reconstructor, depending on the prediction accuracy of the discriminator.

Recently, there have also been some efforts in combining some computer vision tasks and image compression in one framework. In \cite{luo2018deepsic,torfason2018towards}, the authors tried to use the feature maps from learning-based image compression to help other tasks such as image classification and semantic segmentation although the results from other tasks were not used to help the compression part. In \cite{Agustsson18-2}, a segmentation map-based image synthesis model was proposed, which targeted extremely low bit rates ($<$ 0.1 bits/pixel), and used synthesized images for non-important regions. 

\section{Deep-semantic Segmentation-based Layered Image Compression (DSSLIC)}
\label{Proposed Model}

In this section, the proposed semantic segmentation-based layered image compression approach is described. The architecture of the codec and the corresponding deep networks used in the codec are first presented. The loss functions used for training the model are then formulated and explained.

\subsection{DSSLIC Codec Framework}
\label{Codec Architecture}

The overall framework of the DSSLIC codec is shown in Fig. \ref{fig:framework}. The encoder includes three deep learning networks: $SegNet$, $CompNet$, and $FineNet$. The semantic segmentation map $s$ of the input image $x$ is first obtained using $SegNet$. In this paper, a pre-trained PSPNet proposed in \cite{zhao2017pyramid} is used as $SegNet$. The segmentation map is encoded to serve as side information to $CompNet$ for generating a low-dimensional version $c$ of the original image. In this paper, both $s$ and $c$ are losslessly encoded using the FLIF codec \cite{sneyers2016flif}, which is a state-of-the-art lossless image codec.

Given the segmentation map $s$ and compact image $c$, the $RecNet$ part tries to obtain a high-quality reconstruction of the input image. Inside the $RecNet$, the compact image $c$ is first upsampled, which, together with the segmentation map $s$, is fed into a $FineNet$. Note that although GAN-based synthesized images from segmentation maps are visually appealing, their details can be quite different from the original images. To minimize the distortion of the synthesized images, we modify the existing segmentation-based synthesis framework in \cite{wang2017high} and add the upsampled version of the compact image $c$ as an additional input. Besides, $FineNet$ is trained to learn the missing fine information of the upsampled version of $c$ with respect to the input image. This is easier to control the output of the GAN network. After adding the upsampled version of $c$ and the $FineNet$'s output $f$, we get a better estimate of the input. 

In our scheme, if the $SegNet$ fails to assign any label to an area, the $FineNet$ will ignore the semantic input and only reconstruct the image from $c$, which can still get good results. Therefore, our scheme is applicable to all general images. The residual $r$ between the input and the estimate is then obtained and encoded by a lossy codec. In order to deal with negative values, the residual image $r$ is rescaled to [0, 255] with min-max normalization before encoding. The min and max values are also sent to decoder for inverse scaling. In this paper, the H.265/HEVC intra coding-based BPG codec is used \cite{bellard2017bpg}, which is state-of-the-art in lossy coding. 

As a result, in our scheme, the segmentation map $s$ serves as the base layer, and the compact image $c$ and the residual $r$ are respectively the first and second enhancement layers.

At the decoder side, the segmentation map and compact representation are decoded to be used by $RecNet$ to get an estimate of the input image. The output of $RecNet$ is then added to the decoded residual image to get the final reconstruction of the image $\tilde{x}$. The pseudo code of the encoding and decoding procedures is given in Algorithm \ref{alg1}. 

\begin{algorithm}
\caption{DSSLIC Codec}
\label{alg1}
\begin{algorithmic}
\Procedure{Encode}{$x$}
\State $s \gets SegNet(x)$ 
\State $\triangleright$ encode $s$ (1st enhancement layer)
\State $c \gets CompNet(x,s)$ 
\State $\triangleright$ encode $c$ (base layer)
\State $x' \gets RecNet(s,c)$
\State $r  \gets x-x'$ 
\State $min,max \gets Min(r), Max(r)$ 
\State $r  \gets \frac{r-min}{(max-min)}*255$
\State $\triangleright$ encode $r$ (2nd enhancement layer)
\EndProcedure
\end{algorithmic}
\columnbreak
\begin{algorithmic}
\Procedure{Decode}{$s,c,r,min,max$}
\State $x' \gets RecNet(s,c)$
\State $r  \gets \frac{r*(max-min)}{255}+min$
\State $\tilde{x} \gets x' + r$
\EndProcedure
\end{algorithmic}
\begin{algorithmic}
\Function{RecNet}{$s,c$}
\State $c' \gets upsample(c)$
\State $f \gets FineNet(s,c')$
\State $x' \gets c'+f$
\State \Return $x'$
\EndFunction
\end{algorithmic}
\end{algorithm}

\subsection{Network Architecture}

The architectures of the $CompNet$ (proposed in this work) and $FineNet$ (modified from \cite{wang2017high}) networks are defined as follows: 

\begin{itemize}
\item
CompNet:
$c_{64}, d_{128}, d_{256}, d_{512}, c_{3}, tanh$
\item
FineNet:
\\
$c_{64}, d_{128}, d_{256}, d_{512},9\times r_{512}, u_{256}, u_{128}, u_{64}, c_{3}, tanh$
\end{itemize}
where 
\begin{itemize}
\item
$c_k$: 7$\times$7 convolution layers (with $k$ filters and stride 1) followed by instance normalization and ReLU.
\item
$d_k$: 3$\times$3 convolution layers (with $k$ filters and stride 1) followed by instance normalization and ReLU.
\item
$r_k$: a residual block containing reflection padding and two 3$\times$3 convolution layers (with $k$ filters) followed by instance normalization.
\item
$u_k$: 3$\times$3 fractional-strided-convolution layers (with $k$ filters and stride $\frac{1}{2}$) followed by instance normalization and ReLU.
\end{itemize}


Inspired by \cite{wang2017high}, for the adversarial training of the proposed model, two discriminators denoted by $D_1$ and $D_2$ operating at two different image scales are used in this work. $D_1$ operates at the original scale and has a more global view of the image. Thus, the generator can be guided to synthesize fine details in the image. On the other hand, $D_2$ operates with 2$\times$ down-sampled images, leading to coarse information in the synthesized image. Both discriminators have the following architecture:
\begin{itemize}
\item
$C_{64}, C_{128}, C_{256}, C_{512}$
\end{itemize}
where $C_k$ denotes 4$\times$4 convolution layers with $k$ filters and stride 2 followed by instance normalization and LeakyReLU. In order to produce a 1-D output, a convolution layer with 1 filter is utilized after the last layer of the discriminator.


\subsection{Formulation and Objective Functions}
\label{Formulation and Objective}

Let $x \in \mathbb{R}^{h\times w\times k}$ be the original image, the corresponding semantic segmentation map $s \in \mathbb{Z}^{h\times w}$ and the compact representation $c \in \mathbb{R}^{\frac{h}{\alpha} \times \frac{w}{\alpha} \times k}$ are generated as follows:
\begin{equation}
s=SegNet(x), c=CompNet(s,x),
\end{equation}
Conditioned on $s$ and the upscaled $c$, denoted by $c' \in \mathbb{R}^{h\times w\times k}$, $FineNet$ (our GAN generator) reconstructs the fine information image, denoted by $f \in \mathbb{R}^{h\times w\times k}$, which is then added to $c'$ to get the estimate of the input:
\begin{equation}
x'=c'+f, \textrm{   where   } f=FineNet(s,c').
\end{equation}

The error between $x$ and $x'$ is measured using a combination of different losses including $\mathcal{L}_{1}$, $\mathcal{L}_{SSIM}$, $\mathcal{L}_{DIS}$, $\mathcal{L}_{VGG}$, and GAN losses. The L1-norm loss (least absolute errors) is defined as:
\begin{equation}
\mathcal{L}_{1} = 2\lambda {\lVert x-x'\rVert}_1.
\end{equation}

It has been shown that combining pixel-wise losses such as $\mathcal{L}_{1}$ with SSIM loss can significantly improve the perceptual quality of the reconstructed images \cite{zhao2017loss}. As a result, we also utilize the SSIM loss in our work, which is defined as
\begin{equation}
\mathcal{L}_{SSIM} = - I(x,x').C(x,x').S(x,x'),
\end{equation}
where the three comparison functions luminance $I$, contrast $C$, and structure $S$ are computed as:
\begin{equation}
\begin{split}
I(x,x')=\frac{2\mu_x\mu_{x'}+C_1}{\mu^2_x+\mu^2_{x'}+C_1},
C(x,x')=\frac{2\sigma_x\sigma_{x'}+C_2}{\sigma^2_x+\sigma^2_{x'}+C_2},\\
\begin{flalign}
S(x,x')=\frac{\sigma_{xx'}+C_3}{\sigma_x\sigma_{x'}+C_3},
\end{flalign}
\end{split}
\end{equation} 
where $\mu_x$ and $\mu_{x'}$ are the means of $x$ and $x'$, $\sigma_x$ and $\sigma_{x'}$ are the standard deviations, and $\sigma_{xx'}$ is the correlation coefficient. $C1$, $C2$, and $C3$ are the constants used for numerical stability.


To stabilize the training of the generator and produce natural statistics, two perceptual feature-matching losses based on the discriminator and VGG networks \cite{simonyan2014very} are employed. The discriminator-based loss is calculated as:
\begin{equation}
\mathcal{L}_{DIS} = \lambda \sum_{d=1,2} \sum^{n}_{i=1} \frac{1}{N_i}{\lVert D_d^{(i)}(s,c',x)-D_d^{(i)}(s,c',x') \rVert}_1,
\label{eq:pair-wise feature matching}
\end{equation}
where $D_d^{(i)}$ denotes the features extracted from the $i$-th intermediate layer of the discriminator network $D_d$ (with $n$ layers and $N_i$ number of elements in each layer). Similar to \cite{Santurkar17}, a pre-trained VGG network with $m$ layers and $M_j$ elements in each layer is used to construct the VGG perceptual loss as in below:
\begin{equation}
\mathcal{L}_{VGG} =  \lambda \sum^{m}_{j=1} \frac{1}{M_j}{\lVert V^{(j)}(x)-V^{(j)}(x') \rVert}_1,
\label{eq:vgg loss}
\end{equation}
where $V_j$ represents the features extracted from the $j$-th layer of VGG.

In order to distinguish the real training image $x$ from the reconstructed image $x'$, given $s$ and $c'$, the following objective function is minimized by the discriminator $D_d$:
\begin{equation}
\mathcal{L}_D = - \sum_{d=1,2} (\log D_d(s,c',x) + \log(1 - D_d(s,c',x'))),
\end{equation}
while the generator ($FineNet$ in this work) tries to fool $D_d$ by minimizing $- \sum_{d=1,2} \log D_d(s,c',x')$. The final generator loss including all the reconstruction and perceptual losses is then defined as:
\begin{equation}
\mathcal{L}_G = - \sum_{d=1,2} \log D_d(s,c',x') + \mathcal{L}_{1} + \mathcal{L}_{SSIM} + \mathcal{L}_{DIS} + \mathcal{L}_{VGG}.
\end{equation}

Finally, our goal is to minimize the following hybrid loss function: 
\begin{equation}
\mathcal{L} = \mathcal{L}_D + \mathcal{L}_G.
\end{equation}


\subsection{Training}

The Cityscapes (with 30 semantic labels) \cite{cordts2016cityscapes} and ADE20K (with 150 semantic labels) \cite{zhou2017scene} datasets are used for training the proposed model. For Cityscapes, all the 2974 RGB images (street scenes) in the dataset are used. All images are then rescaled to 512$\times$1024 (i.e., $h=512$, $w=1024$, and $k=3$ for RGB channels). For ADE20K, the images with at least 512 pixels in height or width are used (9272 images in total). All images are rescaled to $h=256$ and $w=256$ to have a fixed size for training. Note that no resizing is needed for the test images since the model can work with any size at the testing time. We set the downsampling factor $\alpha=8$ to get the compact representation of size 64$\times$128$\times$3 for Cityscapes and 32$\times$32$\times$3 for ADE20K. We also consider the weight $\lambda=10$ for $\mathcal{L}_{1}$, $\mathcal{L}_{DIS}$, and $\mathcal{L}_{VGG}$.

All models were jointly trained for 150 epochs with mini-batch stochastic gradient descent (SGD) and a mini-batch sizes of 2 and 8 for Cityscapes and ADE20K, respectively. The Adam solver with learning rate of 0.0002 was used, which is fixed for the first 100 epochs, but gradually decreases to zero for the next 50 epochs. Perceptual feature-matching losses usually guide the generator towards more synthesized textures in the predicted images, which causes a slightly higher pixel-wise reconstruction error, especially in the last epochs. To handle this issue, we did not consider the perceptual $\mathcal{L}_{D}$ and $\mathcal{L}_{VGG}$ losses in the generator loss for the last 50 epochs. All the $SegNet$, $CompNet$, $FineNet$, and the discriminator networks proposed in this work are trained in the RGB domain.


\section{Experiments}
\label{Experiments}

In this section, we compare the performance of the proposed DSSLIC scheme with JPEG, JPEG2000, WebP, and the H.265/HEVC intra coding-based BPG codec \cite{bellard2017bpg}, which is state-of-the-art in lossy image compression. Since the networks are trained for RGB images, we encode all images using RGB (4:4:4) format in different codecs for fair comparison. We use both PSNR and MS-SSIM \cite{wang2003multiscale} as the evaluation metric in this experiment. In this experiment, we encode the RGB components of the residual image $r$ using lossy BPG codec with different quantization values.

\begin{figure*}
\begin{subfigure}[b]{0.485\linewidth}
 \centering
  \centerline{\includegraphics[width=\textwidth]{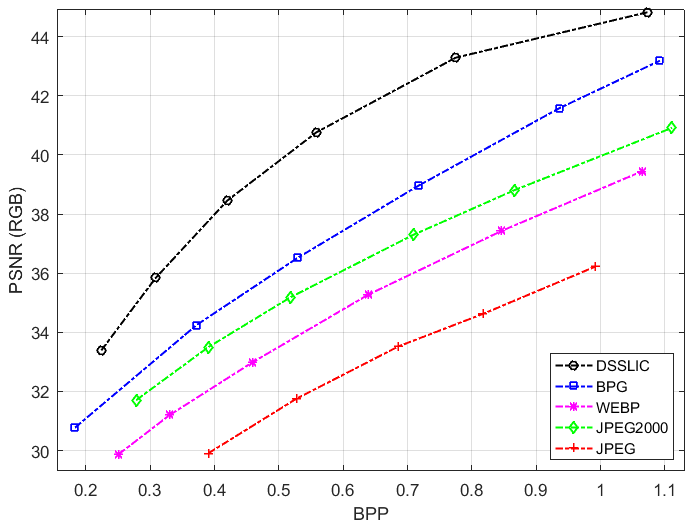}}
\end{subfigure}
\begin{subfigure}[b]{0.485\linewidth}
 \centering
  \centerline{\includegraphics[width=\textwidth]{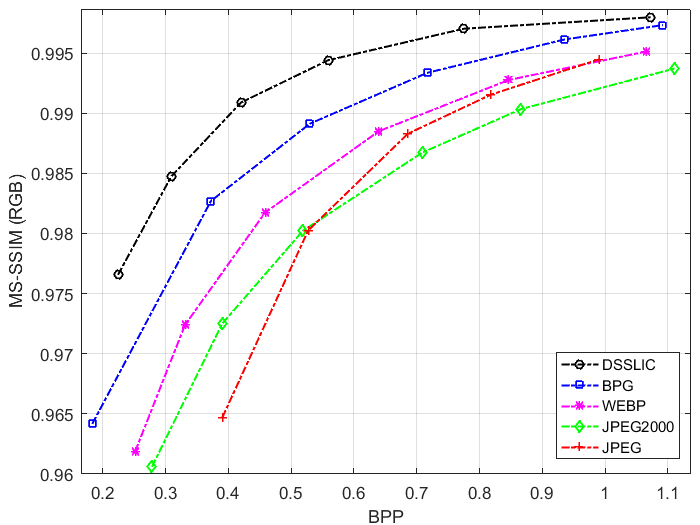}}
\end{subfigure}
\caption{Comparison results on ADE20K test set in terms of PSNR (left) and MS-SSIM (right) vs. bpp (bits/pixel/channel). The results are averaged over RGB channels.}
\label{fig:results_ADE}
\end{figure*}

\begin{figure*}
\begin{subfigure}[b]{0.485\linewidth}
 \centering
  \centerline{\includegraphics[width=\textwidth]{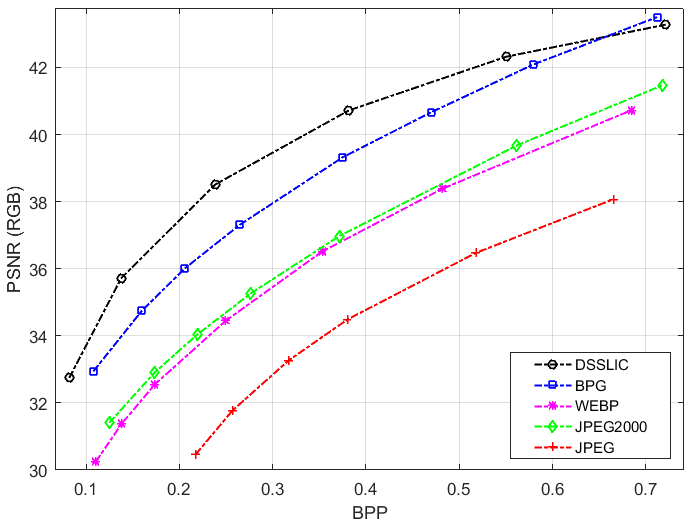}}
\end{subfigure}
\begin{subfigure}[b]{0.485\linewidth}
 \centering
  \centerline{\includegraphics[width=\textwidth]{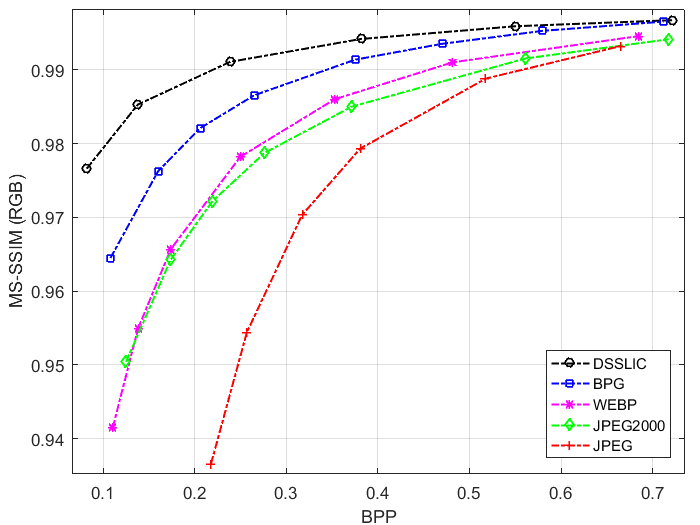}}
\end{subfigure}
\caption{Comparison results on Cityscapes test set.}
\label{fig:results_City}
\end{figure*}

The results of the ADE20K and Cityscapes test sets are given in Figures \ref{fig:results_ADE} and \ref{fig:results_City}. The results are averaged over 50 random test images not included in the training set. As shown in the figures, our method gives better PSNR and MS-SSIM than BPG, especially when the bit rate is less than $\approx0.9$ bits/pixel/channel (bpp for short) on ADE20K and less than $\approx0.5$ bpp on Cityscapes. In particular, the average PSNR gain is more than 2dB for the ADE20k test set when the bit rate is between 0.4-0.7 bpp.

\begin{figure*}
\begin{subfigure}[b]{0.485\linewidth}
 \centering
  \centerline{\includegraphics[width=\textwidth]{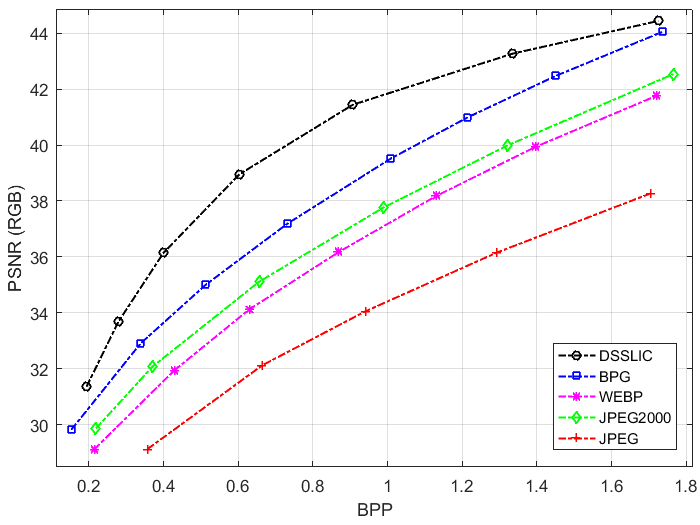}}
\end{subfigure}
\begin{subfigure}[b]{0.485\linewidth}
 \centering
  \centerline{\includegraphics[width=\textwidth]{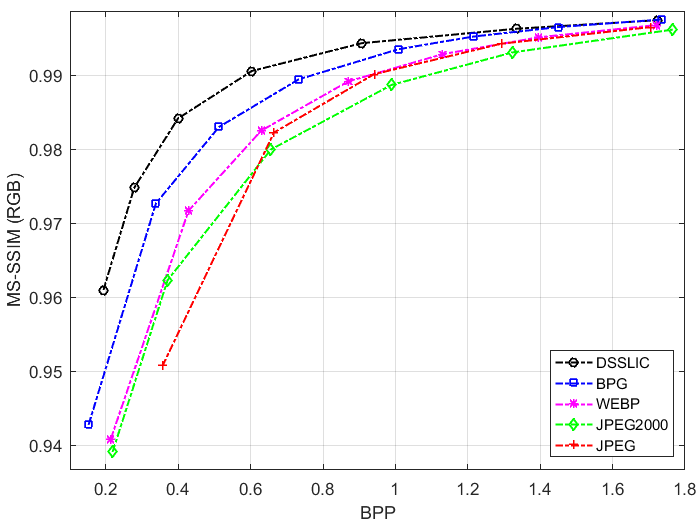}}
\end{subfigure}
\caption{Comparison results on Kodak image set.}
\label{fig:results_Kodak}
\end{figure*}

To demonstrate the generalization capability of the scheme, the ADE20K-trained model is also applied to the classical Kodak dataset (including 24 test images). The average results of the Kodak dataset are illustrated in Figure \ref{fig:results_Kodak}. For this experiment, the model trained on the ADE20K dataset is used. It is shown that when the bit rate is less than about 1.4 bpp, our scheme achieves better results than other codecs in both PSNR and MS-SSIM. For example, the average gain is about 2 dB between 0.4-0.8 bpp. This is quite promising since the proposed scheme can still be improved in many ways. This also shows that our method generalizes very well when the training and testing images are from different datasets. The average $RecNet$ decoding time for Kodak images on CPU and GPU are $\approx$44s and $\approx$0.013s, respectively.

\begin{figure*}
\centering
\begin{subfigure}[b]{.275\textwidth}
 \centering
  \centerline{\includegraphics[width=\textwidth]{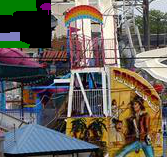}}
  \subcaption{\scriptsize Original (with segmentation map)}
\end{subfigure}
\begin{subfigure}[b]{.275\textwidth}
 \centering
  \centerline{\includegraphics[width=\textwidth]{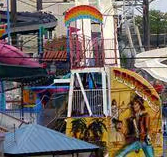}}
  \subcaption{\scriptsize DSSLIC (ours) (0.59 bpp, 31.38 dB, 0.988)}
\end{subfigure}
\begin{subfigure}[b]{.275\textwidth}
 \centering
  \centerline{\includegraphics[width=\textwidth]{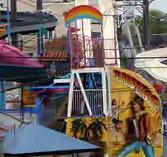}}
  \subcaption{\scriptsize BPG (0.59 bpp, 27.31 dB, 0.984)}
\end{subfigure}
\\
\begin{subfigure}[b]{.275\textwidth}
 \centering
  \centerline{\includegraphics[width=\textwidth]{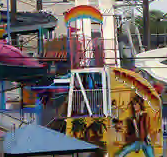}}
  \subcaption{\scriptsize WebP (0.60 bpp, 25.43 dB, 0.979)}
\end{subfigure}
\begin{subfigure}[b]{.275\textwidth}
 \centering
  \centerline{\includegraphics[width=\textwidth]{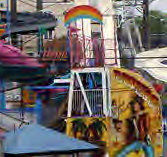}}
  \subcaption{\scriptsize JPEG2000 (0.60 bpp, 25.12 dB, 0.972)}
\end{subfigure}
\begin{subfigure}[b]{.275\textwidth}
 \centering
  \centerline{\includegraphics[width=\textwidth]{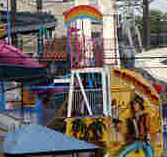}}
  \subcaption{\scriptsize JPEG (0.61 bpp, 23.63 dB, 0.973)}
\end{subfigure}
\caption{ADE20K visual example 1. (bits/pixel/channel, PSNR, MS-SSIM)}
\label{fig:ADE_quan1}
\end{figure*}

\begin{figure*}
\centering
\begin{subfigure}[b]{.275\textwidth}
 \centering
  \centerline{\includegraphics[width=\textwidth]{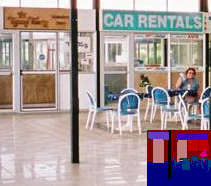}}
  \subcaption{\scriptsize Original (with segmentation map)}
\end{subfigure}
\begin{subfigure}[b]{.275\textwidth}
 \centering
  \centerline{\includegraphics[width=\textwidth]{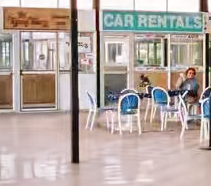}}
  \subcaption{\scriptsize DSSLIC (ours) (0.18 bpp, 30.87 dB, 0.973)}
\end{subfigure}
\begin{subfigure}[b]{.275\textwidth}
 \centering
  \centerline{\includegraphics[width=\textwidth]{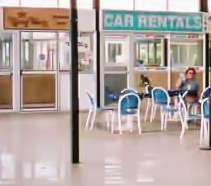}}
  \subcaption{\scriptsize BPG (0.18 bpp, 29.93 dB, 0.965)}
\end{subfigure}
\\
\begin{subfigure}[b]{.275\textwidth}
 \centering
  \centerline{\includegraphics[width=\textwidth]{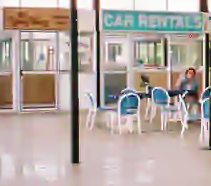}}
  \subcaption{\scriptsize WebP (0.21 bpp, 27.48 dB, 0.956)}
\end{subfigure}
\begin{subfigure}[b]{.275\textwidth}
 \centering
  \centerline{\includegraphics[width=\textwidth]{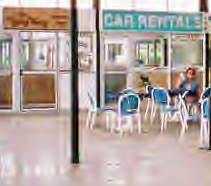}}
  \subcaption{\scriptsize JPEG2000 (0.21 bpp, 27.13 dB, 0.946)}
\end{subfigure}
\begin{subfigure}[b]{.275\textwidth}
 \centering
  \centerline{\includegraphics[width=\textwidth]{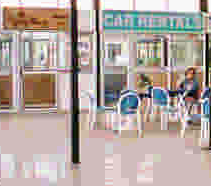}}
  \subcaption{\scriptsize JPEG (0.23 bpp, 24.61 dB, 0.907)}
\end{subfigure}
\caption{ADE20K visual example 2. (bits/pixel/channel, PSNR, MS-SSIM)}
\label{fig:ADE_quan2}
\end{figure*}

\begin{figure*}
\centering
\begin{subfigure}[b]{.28\textwidth}
 \centering
  \centerline{\includegraphics[width=\textwidth]{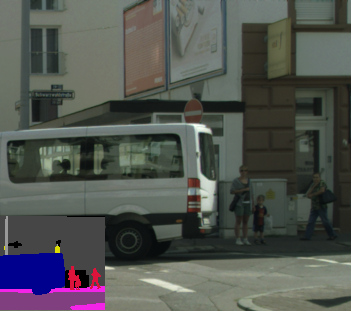}}
  \subcaption{\scriptsize Original (with segmentation map)}
\end{subfigure}
\begin{subfigure}[b]{.28\textwidth}
 \centering
  \centerline{\includegraphics[width=\textwidth]{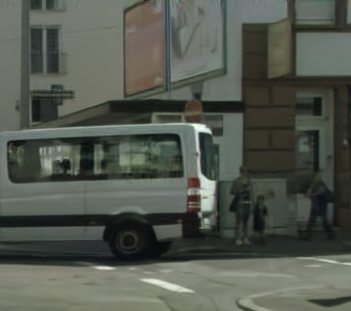}}
  \subcaption{\scriptsize DSSLIC (ours) (0.12 bpp, 34.72 dB, 0.987)}
\end{subfigure}
\begin{subfigure}[b]{.28\textwidth}
 \centering
  \centerline{\includegraphics[width=\textwidth]{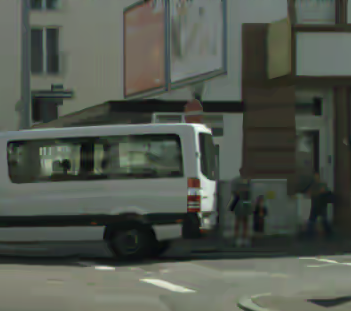}}
  \subcaption{\scriptsize BPG (0.12 bpp, 33.49 dB, 0.976)}
\end{subfigure}
\\
\begin{subfigure}[b]{.28\textwidth}
 \centering
  \centerline{\includegraphics[width=\textwidth]{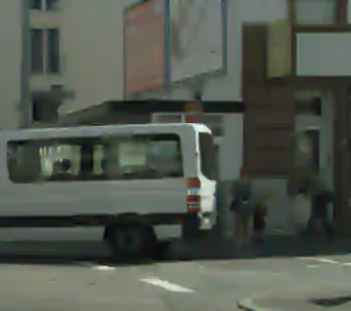}}
  \subcaption{\scriptsize WebP (0.12 bpp, 29.89 dB, 0.954)}
\end{subfigure}
\begin{subfigure}[b]{.28\textwidth}
 \centering
  \centerline{\includegraphics[width=\textwidth]{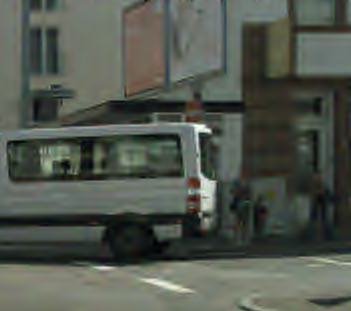}}
  \subcaption{\scriptsize JPEG2000 (0.12 bpp, 30.26 dB, 0.953)}
\end{subfigure}
\begin{subfigure}[b]{.28\textwidth}
 \centering
  \centerline{\includegraphics[width=\textwidth]{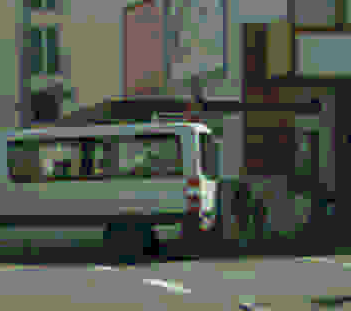}}
  \subcaption{\scriptsize JPEG (0.14 bpp, 24.81 dB, 0.852)}
\end{subfigure}
\caption{Cityscapes visual example 1. (bits/pixel/channel, PSNR, MS-SSIM)}
\label{fig:quan1}
\end{figure*}

\begin{figure*}
\centering
\begin{subfigure}[b]{.28\textwidth}
 \centering
  \centerline{\includegraphics[width=\textwidth]{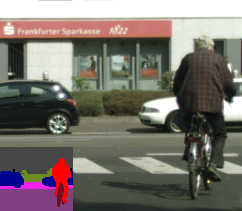}}
  \subcaption{\scriptsize Original (with segmentation map)}
\end{subfigure}
\begin{subfigure}[b]{.28\textwidth}
 \centering
  \centerline{\includegraphics[width=\textwidth]{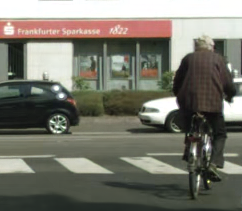}}
  \subcaption{\scriptsize DSSLIC (ours) (0.21 bpp, 37.30 dB, 0.994)}
\end{subfigure}
\begin{subfigure}[b]{.28\textwidth}
 \centering
  \centerline{\includegraphics[width=\textwidth]{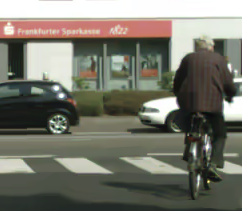}}
  \subcaption{\scriptsize BPG (0.21 bpp, 36.32 dB, 0.992)}
\end{subfigure}
\\
\begin{subfigure}[b]{.28\textwidth}
 \centering
  \centerline{\includegraphics[width=\textwidth]{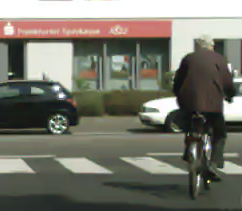}}
  \subcaption{\scriptsize WebP (0.21 bpp, 33.06 dB, 0.986)}
\end{subfigure}
\begin{subfigure}[b]{.28\textwidth}
 \centering
  \centerline{\includegraphics[width=\textwidth]{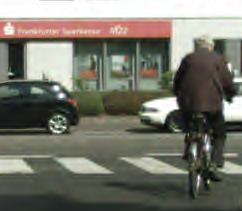}}
  \subcaption{\scriptsize JPEG2000 (0.21 bpp, 32.26 dB, 0.980)}
\end{subfigure}
\begin{subfigure}[b]{.28\textwidth}
 \centering
  \centerline{\includegraphics[width=\textwidth]{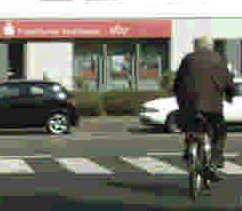}}
  \subcaption{\scriptsize JPEG (0.21 bpp, 28.10 dB, 0.947)}
\end{subfigure}
\caption{Cityscapes visual example 2. (bits/pixel/channel, PSNR, MS-SSIM)}
\label{fig:quan2}
\end{figure*}

\begin{figure*}
\centering
\begin{subfigure}[b]{.29\textwidth}
 \centering
  \centerline{\includegraphics[width=\textwidth]{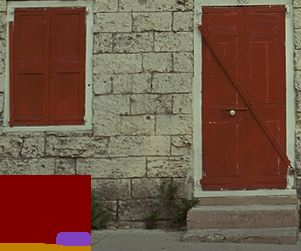}}
  \subcaption{\scriptsize Original (with segmentation map)}
\end{subfigure}
\begin{subfigure}[b]{.29\textwidth}
 \centering
  \centerline{\includegraphics[width=\textwidth]{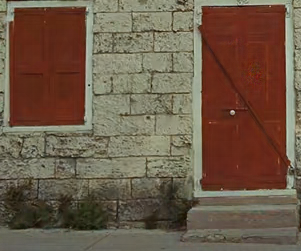}}
  \subcaption{\scriptsize DSSLIC (ours) (0.48 bpp, 33.26 dB, 0.984)}
\end{subfigure}
\begin{subfigure}[b]{.29\textwidth}
 \centering
  \centerline{\includegraphics[width=\textwidth]{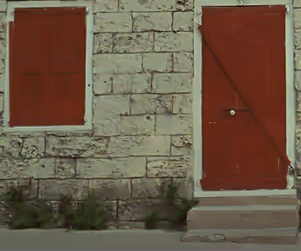}}
  \subcaption{\scriptsize BPG (0.49 bpp, 29.56 dB, 0.971)}
\end{subfigure}
\\
\begin{subfigure}[b]{.29\textwidth}
 \centering
  \centerline{\includegraphics[width=\textwidth]{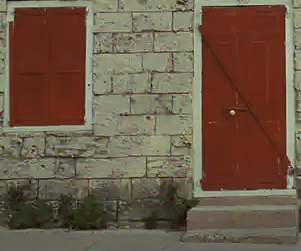}}
  \subcaption{\scriptsize WebP (0.50 bpp, 27.91 dB, 0.961)}
\end{subfigure}
\begin{subfigure}[b]{.29\textwidth}
 \centering
  \centerline{\includegraphics[width=\textwidth]{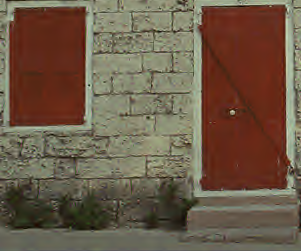}}
  \subcaption{\scriptsize JPEG2000 (0.49 bpp, 27.82 dB, 0.953)}
\end{subfigure}
\begin{subfigure}[b]{.29\textwidth}
 \centering
  \centerline{\includegraphics[width=\textwidth]{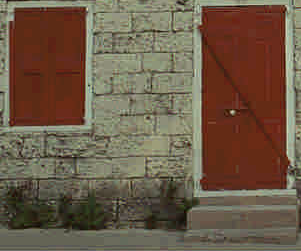}}
  \subcaption{\scriptsize JPEG (0.50 bpp, 26.34 dB, 953)}
\end{subfigure}
\caption{Kodak visual example 1. (bits/pixel/channel, PSNR, MS-SSIM)}
\label{fig:kodak_quan1}
\end{figure*}

\begin{figure*}
\centering
\begin{subfigure}[b]{.29\textwidth}
 \centering
  \centerline{\includegraphics[width=\textwidth]{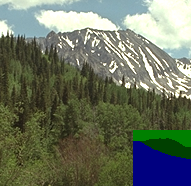}}
  \subcaption{\scriptsize Original (with segmentation map)}
\end{subfigure}
\begin{subfigure}[b]{.29\textwidth}
 \centering
  \centerline{\includegraphics[width=\textwidth]{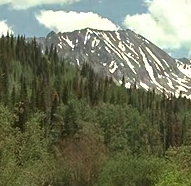}}
  \subcaption{\scriptsize DSSLIC (ours) (0.69 bpp, 32.54 dB, 0.982)}
\end{subfigure}
\begin{subfigure}[b]{.29\textwidth}
 \centering
  \centerline{\includegraphics[width=\textwidth]{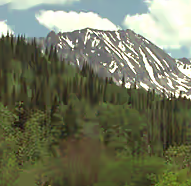}}
  \subcaption{\scriptsize BPG (0.71 bpp, 27.86 dB, 0.957)}
\end{subfigure}
\\
\begin{subfigure}[b]{.29\textwidth}
 \centering
  \centerline{\includegraphics[width=\textwidth]{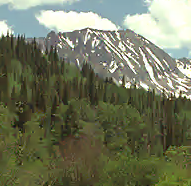}}
  \subcaption{\scriptsize WebP (0.71 bpp, 26.01 dB, 0.952)}
\end{subfigure}
\begin{subfigure}[b]{.29\textwidth}
 \centering
  \centerline{\includegraphics[width=\textwidth]{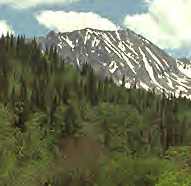}}
  \subcaption{\scriptsize JPEG2000 (0.71 bpp, 26.71 dB, 0.942)}
\end{subfigure}
\begin{subfigure}[b]{.29\textwidth}
 \centering
  \centerline{\includegraphics[width=\textwidth]{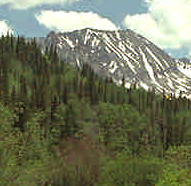}}
  \subcaption{\scriptsize JPEG (0.72 bpp, 24.77 dB, 0.958)}
\end{subfigure}
\caption{Kodak visual example 2. (bits/pixel/channel, PSNR, MS-SSIM)}
\label{fig:kodak_quan2}
\end{figure*}

\begin{figure*}
\centering
\begin{subfigure}[b]{.30\textwidth}
 \centering
  \centerline{\includegraphics[width=\textwidth]{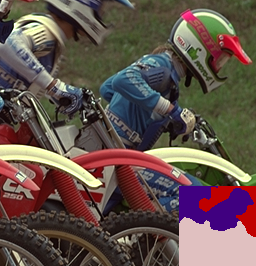}}
  \subcaption{\scriptsize Original (with segmentation map)}
\end{subfigure}
\begin{subfigure}[b]{.30\textwidth}
 \centering
  \centerline{\includegraphics[width=\textwidth]{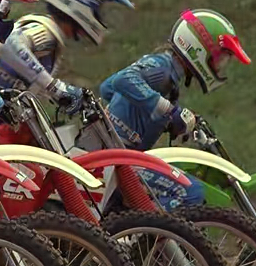}}
  \subcaption{\scriptsize DSSLIC (ours) (0.40 bpp, 30.68 dB, 0.978)}
\end{subfigure}
\begin{subfigure}[b]{.30\textwidth}
 \centering
  \centerline{\includegraphics[width=\textwidth]{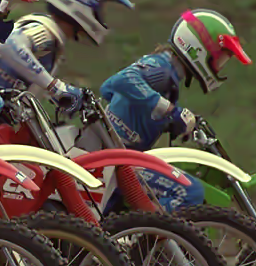}}
  \subcaption{\scriptsize BPG (0.40 bpp, 28.72 dB, 0.969)}
\end{subfigure}
\\
\begin{subfigure}[b]{.30\textwidth}
 \centering
  \centerline{\includegraphics[width=\textwidth]{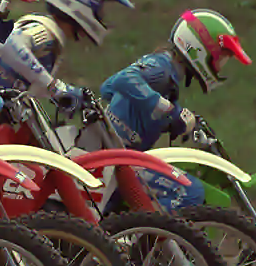}}
  \subcaption{\scriptsize WebP (0.41 bpp, 26.63 dB, 0.957)}
\end{subfigure}
\begin{subfigure}[b]{.30\textwidth}
 \centering
  \centerline{\includegraphics[width=\textwidth]{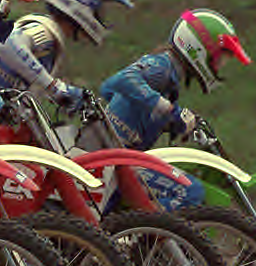}}
  \subcaption{\scriptsize JPEG2000 (0.41 bpp, 26.49 dB, 0.949)}
\end{subfigure}
\begin{subfigure}[b]{.30\textwidth}
 \centering
  \centerline{\includegraphics[width=\textwidth]{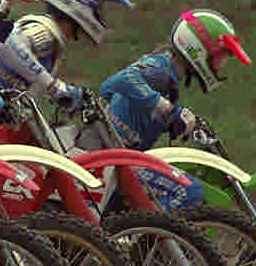}}
  \subcaption{\scriptsize JPEG (0.41 bpp, 24.63 dB, 0.941)}
\end{subfigure}
\caption{Kodak visual example 3. (bits/pixel/channel, PSNR, MS-SSIM)}
\label{fig:kodak_quan3}
\end{figure*}

Some visual examples from ADE20K, Cityscapes, and Kodak test sets are given in Figures \ref{fig:ADE_quan1}-\ref{fig:kodak_quan3}. In order to have a more clear visualization, only some cropped parts of the reconstructed images are shown in these examples. As seen in all examples, JPEG has poor performance due to the blocking artifacts. Some artifacts are also seen on JPEG2000 results. Although WebP provides higher quality results than JPEG2000, the images are blurred in some areas. The images encoded using BPG are smoother, but the fine structures are also missing in some areas.

\begin{figure*}
\begin{minipage}[b]{0.19\linewidth}
 \centering
  \centerline{\includegraphics[width=\textwidth]{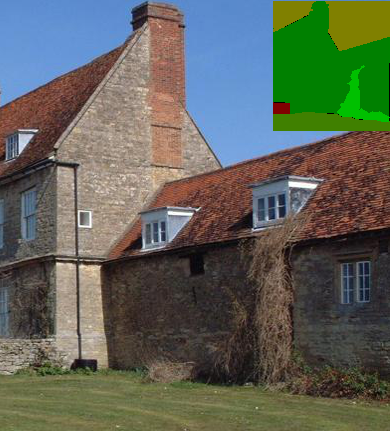}}
  \subcaption{Original\\PSNR, MS-SSIM}
\end{minipage}
\begin{minipage}[b]{0.19\linewidth}
 \centering
  \centerline{\includegraphics[width=\textwidth]{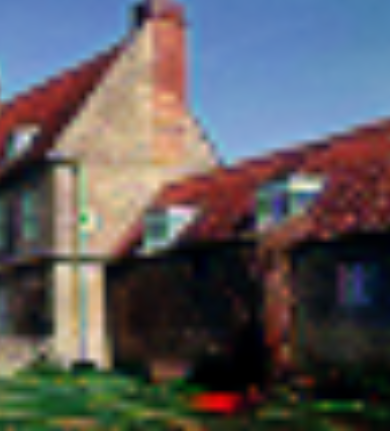}}
  \subcaption{upComp\\18.01 dB, 0.73}
\end{minipage}
\begin{minipage}[b]{0.19\linewidth}
 \centering
  \centerline{\includegraphics[width=\textwidth]{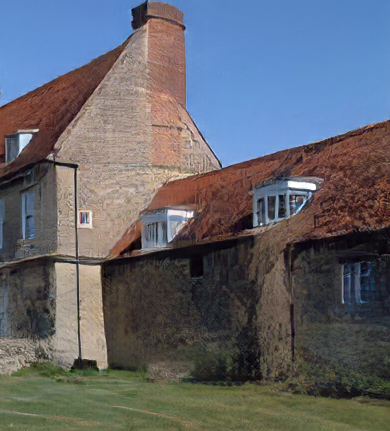}}
  \subcaption{synth\\23.68 dB, 0.84}
\end{minipage}
\begin{minipage}[b]{0.19\linewidth}
 \centering
  \centerline{\includegraphics[width=\textwidth]{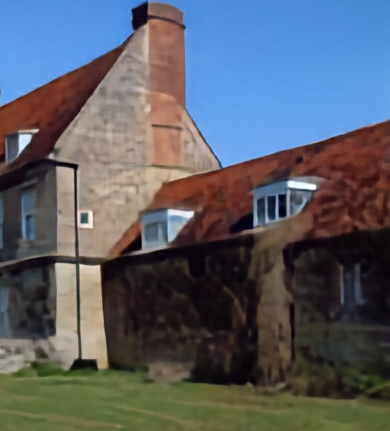}}
  \subcaption{noSeg\\22.18 dB, 0.86}
\end{minipage}
\begin{minipage}[b]{0.19\linewidth}
 \centering
  \centerline{\includegraphics[width=\textwidth]{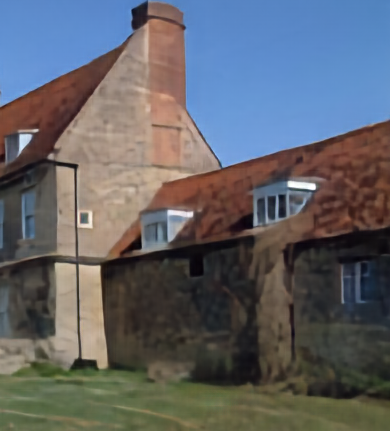}}
  \subcaption{withSeg\\25.09 dB, 0.88}
\end{minipage}
\caption{Visual comparison of different scenarios at 0.08 BPP.}
\label{fig:ablation}
\end{figure*}

\begin{table*}
\caption{Results of different scenarios (without BPG-based residual coding).}
\centering
\begin{tabular}{|l|c|c|c|c||c|c|c|c|} 
\cline{2-9}
\multicolumn{1}{c|}{} & \multicolumn{4}{c||}{\textbf{ADE20K} }  & \multicolumn{4}{c|}{\textbf{Kodak}}  \\ 
\cline{2-9}
\multicolumn{1}{c|}{} & \textbf{upComp}  & \textbf{synth} & \textbf{noSeg} & \textbf{withSeg}  & \textbf{upComp} & \textbf{synth}  & \textbf{noSeg}  & \textbf{withSeg}   \\ 
\hline
\textbf{BPP}        &  0.095   & 0.092  & 0.08   & 0.095        &        0.087 & 0.088       & 0.080           & 0.087           \\
\textbf{PSNR}       &  17.50   & 21.91  & 22.24   & 23.11       &          17.77    & 20.97   & 21.46           & 21.91           \\
\textbf{MS-SSIM}    &  0.759   & 0.887  & 0.905   & 0.914       &   0.738  & 0.858      & 0.887           & 0.891           \\
\hline
\end{tabular}
\label{tbl:results}
\end{table*}

Figure \ref{fig:ablation} and Table \ref{tbl:results} report some ablation studies of different configurations, all are obtained without using the BPG-based residual coding, including: 
\textbf{upComp}: the results are obtained without considering the $FineNet$ network in the pipeline, i.e., $x'=c'$ (the upsampled compact image only);
\textbf{noSeg}: the segmentation maps are not considered in neither $CompNet$ nor $FineNet$ networks, i.e., $x'=c'+f$ where $c'$ is the upsampled version of $c=CompNet(x)$, and $f=FineNet(c')$;
\textbf{withSeg}: all the DSSLIC components shown in Figure \ref{fig:framework} are used in this configuration (except BPG-based residual coding);
\textbf{synth}:, the settings in this configuration is the same as withSeg except that the perceptual losses $\mathcal{L}_{VGG}$ and $\mathcal{L}_{DIS}$ are considered in all training epochs. The poor performance of using only the upsampled compact images in \textbf{upComp} shows the importance of $FN$ in predicting the missing fine information, which is also visually obvious in Figure \ref{fig:ablation}. Considering perceptual losses in all training epochs (\textbf{synth}) leads to sharper and perceptually more natural images, but the PSNR is much lower. The results with segmentation maps (\textbf{withSeg}) provide slightly better PSNR than \textbf{noSeg} although the visual gain is more pronounced, e.g., the dark wall in Figure \ref{fig:ablation}. 


\section{Conclusion}
\label{Conclusion}

In this paper, we proposed a deep semantic segmentation-based layered image compression (DSSLIC) framework in which the semantic segmentation map of the input image was used to synthesize the image, and the residual was encoded as an enhancement layer in the bit-stream. 

Experimental results showed that the proposed framework outperforms the H.265/HEVC-based BPG and the other standard codecs in both PSNR and MS-SSIM metrics in RGB (4:4:4) domain. In addition, since semantic segmentation map is included in the bit-stream, the proposed scheme can facilitate many other tasks such as image search and object-based adaptive image compression. 

The proposed scheme opens up many future topics, for example, improving its high-rate performance, modifying the scheme for YUV-coded images, and applying the framework for other tasks.


{\small
\bibliographystyle{ieee}
\bibliography{main}
}

\end{document}